\documentclass[letterpaper]{article}
\usepackage[preprint]{aaai2027}

\usepackage[hyphens]{url}
\usepackage{graphicx}
\usepackage{natbib}
\usepackage{caption}
\usepackage{algorithm}
\usepackage{algorithmic}

\usepackage{newfloat}
\usepackage{listings}
\DeclareCaptionStyle{ruled}{labelfont=normalfont,labelsep=colon,strut=off}
\floatstyle{ruled}
\newfloat{listing}{tb}{lst}{}
\floatname{listing}{Listing}

\usepackage{booktabs}
\usepackage{amsmath}
\usepackage{amsfonts}

\title{Training and Evaluating Ethical Reinforcement Learning Agents \\ on Per-Episode Distributions}
\author{
    Prabhjyot Singh\textsuperscript{\rm 1}\corresponding,
    Majid Ghasemi\textsuperscript{\rm 1},
    Mark Crowley\textsuperscript{\rm 1}
}
\affiliations{
    \textsuperscript{\rm 1}Department of Electrical and Computer Engineering, University of Waterloo\\
    {prabhjyot.singh, majid.ghasemi, mark.crowley}@uwaterloo.ca\\
    \textit{Under review}
}

\begin{document}

\maketitle

\begin{abstract}
Reinforcement Learning (RL) agents trained on a single reward signal exploit the gap between the designed reward and the intended behavior. This is particularly a problem when we are trying to imbue ethical behavior into RL agents. An agent can look ethical \emph{on average} while concentrating its violations in a few bad episodes, and a creature in the environment harmed in one episode is not restored by good conduct in another. We compare four ways of training ethical behavior in Craftax, an open-ended survival benchmark. The four are: scalar penalties with termination, a linear multi-objective weight sweep, an adaptive Lagrangian constraint, and a non-compensatory utility optimized per episode under the Expected Scalarized Returns (ESR) criterion. All are evaluated under a single detector-based protocol that counts every violation in every episode without censoring. On the frontier of mean return against mean violation rate, the four methods are indistinguishable; per episode they separate sharply. At matched mean return, the ESR agent holds its stated budget of one violation in effectively every episode (worst-decile $1.04 \pm 0.07$ violations), the Lagrangian leaks past the same budget ($1.14 \pm 0.03$), and the weight sweep's worst episodes double it ($2.20 \pm 0.20$). An observation-augmentation control attributes the separation to the training objective rather than to what the agent observes, and the per-episode guarantee costs nothing on the mean frontier. When ethical violations do not average away across episodes, we argue both training and evaluation must target the per-episode distribution rather than the mean. 
\end{abstract}

\section{Introduction}

Reinforcement Learning (RL) agents that chase a single reward signal often find ways to score well that their designers never wanted. This is specification gaming, or reward hacking \citep{krakovna2020specification, skalse2022defining}, and it arises whenever the reward we can write down is only a rough stand-in for what we actually care about \citep{amodei2016concrete}. The damage is mild in small closed tasks but severe in open-ended worlds, the setting that most resembles real deployment. Craftax \citep{matthews2024craftax} is one such world, a fast JAX survival game in which an agent gathers resources, climbs a technology tree, and survives across procedurally generated levels. Its speed is a gift for research, but it also lets a misaligned agent practice harmful shortcuts at enormous scale. Rewarded only for progress, the agent will strip a forest bare or kill a harmless creature for a few more points.

Ethical training fails in a particular way. An agent whose violations are rare on average may still concentrate them in a handful of bad episodes, and harm done in one episode is not undone by good conduct in another. Mean violation rate, the standard target for both training and evaluation, cannot tell a uniformly almost-clean agent apart from one that is perfect most of the time and catastrophic occasionally. For the  goal of learning ethically aligned behavior, this hidden tail is unacceptable.

We address this challenge with two design choices, the first being the \emph{shape} of the trade-off. Rather than a hand-tuned linear weighting, we formalize the ethical requirement as a non-compensatory utility, a thresholded-lexicographic rule under which task return counts only while an episode stays within a stated violation budget. As a secondary variant we \emph{distill} the same stated ordering into a differentiable function via Bradley--Terry preference learning \citep{wirth2017survey, christiano2017deep}. The labels come from our own rule, so this distills a stated ethic rather than learning ethics from behavior. The second choice is \emph{when} the trade-off applies. A utility can be applied after averaging over episodes, letting a bad episode be offset by good ones, or inside each episode \citep{roijers2013survey, hayes2022practical}. For ethical behavior we take the second view, so violations are counted per episode and never averaged away.

Measurement needs the same care as objective design: if we collapse violations to a single mean, we reintroduce the same cross-episode compensation that makes rare catastrophic episodes invisible. We score every trained agent, agnostic of its training method, in one shared detector-only environment with uncensored per-step counts and report the per-episode violation \emph{distribution}, its standard deviation, the probability of any violation, and the worst-decile mean (CVaR). Across three Craftax dilemmas, we compare four training methods under this protocol: scalar penalties with termination, a linear weight sweep, an adaptive Lagrangian constraint, and the per-episode non-compensatory objective. The objective \emph{removes cross-episode compensation}, so it cannot be satisfied by being ethical on average, but it does not guarantee a clean episode, since policies and environments stay stochastic.

\newpage

Our contributions are as follows:
\begin{itemize}
    \item A confound-controlled  comparison of four training methods for ethical RL behavior, with controls separating the objective (ESR vs.\ SER), the observation (SER$+R_{\mathrm{acc}}$), the utility source (stated vs.\ distilled), and the discounting approximation ($\gamma{=}1$).

    \item An evaluation methodology, a shared detector-only environment, uncensored counts, and per-episode distributional metrics, with the argument that mean rates cannot distinguish ``ethical'' from ``ethical on average''.

    \item A reward-isolated ethical-dilemma benchmark in Craftax, three dilemmas $\times$ three enforcement mechanisms, released with code and training pipeline.

    \item We find that the four methods are indistinguishable based on the mean frontier approach yet cleanly separated per episode. The non-compensatory objective holds its budget in effectively every episode at no cost in mean return, and the observation control attributes the per-episode budget ceiling to the objective rather than to what the agent sees.
    
\end{itemize}

\section{Related Work}

\paragraph{Specification gaming.}
Reward hacking, an agent exploiting the gap between a proxy reward and the intended one, is a well known safety concern \citep{krakovna2020specification, amodei2016concrete}, made precise by asking when improving a proxy can lower the true reward \citep{skalse2022defining}. We study it in an open-ended world where such shortcuts are plentiful and cheap to repeat.

\paragraph{Multi-Objective Reinforcement Learning (MORL).}
MORL keeps objectives as a vector and applies a utility to act \citep{roijers2013survey, hayes2022practical}, and most systems fix a linear utility with hand-chosen weights. A weight sweep recovers only the convex hull of the achievable front \citep{vamplew2011empirical}, whereas non-linear scalarizations reach more \citep{van2014multi}, and our stated utility is a smoothed thresholded-lexicographic ordering \citep{gabor1998multi}. More important here is \emph{when} the utility applies. Under Scalarized Expected Returns (SER) it follows the expectation, under Expected Scalarized Returns (ESR) it precedes it \citep{roijers2020multi, ruadulescu2020multi}. We take the ESR view, since an ethical violation happens within a single episode. 
Algorithmically we are closest to the ESR actor-critic of \citet{reymond2023actor}, which conditions the policy on the accrued reward to optimize a non-linear utility. Our ESR-PPO is that idea in PPO form, the vehicle rather than the contribution, and what we add is the per-episode ethical budget it encodes and the controlled, distribution-level comparison against the linear alternatives.

\paragraph{Safe and constrained reinforcement learning.}
A separate tradition enforces safety as a constraint. Constrained MDPs cap expected cost \citep{altman2021constrained}, many methods respect such limits \citep{garcia2015comprehensive, achiam2017constrained}, and shielding blocks unsafe actions outright \citep{alshiekh2018safe} but leaves the safety in the machinery rather than the agent \citep{ghasemi2026objective}. 
We do not use this approach, but we do implement a dual-ascent Lagrangian \citep{altman2021constrained, achiam2017constrained, ray2019benchmarking} as a first-class baseline, since an expected-cost constraint is exactly what a per-episode claim must beat. Saute RL \citep{sootla2022saute} augments the state with a cost budget to satisfy a hard cap almost surely, mechanically like our accrued-return augmentation, but we optimize a graded utility over the return vector, not a single cap. 
Risk-sensitive RL controls tails via CVaR or chance constraints \citep{chow2018risk}. We use CVaR to \emph{evaluate} all methods, not to train.

\paragraph{Learning rewards from preferences.}
Preference-based reward learning fits a model to trajectory comparisons \citep{wirth2017survey}, often with a Bradley--Terry likelihood \citep{bradley1952rank, christiano2017deep}. What is learned depends on where the comparisons come from. Most use human raters, whereas we generate them mechanically from a hand-coded ethics-gated rule plus a monotonicity prior. Our fitted utility thus \emph{distills} a stated ordering into a differentiable function, an ablation on whether fitting helps rather than ethics learned from behavior. The pool it is fit to simply reuses agents trained elsewhere in this study.

\paragraph{Machine ethics as multiple objectives.}
Alignment is widely framed as pluralistic \citep{sorensen2024value, graham2013moral}, and \citet{vamplew2018human} argue that ethical, legal, and safety constraints are competing objectives a linear scalar utility cannot serve. Closest to us, \citet{rodriguez2021multi} embed ethics in a multi-objective process but weight it linearly as designers, and MORAL \citep{peschl2021moral} tunes a distribution over linear weights through human queries. We differ by using a non-linear utility optimized under ESR, so each episode is judged on its own, and the non-linearity is what lets a breach resist averaging. We run in Craftax \citep{matthews2024craftax}, an open-ended JAX benchmark in the lineage of Crafter \citep{hafner2021benchmarking} and NetHack \citep{kuttler2020nethack}, and its speed doubles as a tool to stress test various things.

\section{Training Methods}

\subsection{Problem Setting and Optimization Criteria}
We model each dilemma as a multi-objective MDP $\langle \mathcal{S}, \mathcal{A}, P, \mathbf{r}, \gamma \rangle$ with a two-component reward $\mathbf{r}_t = [\, r_{\mathrm{ext},t},\; r_{\mathrm{eth},t} \,]$. Here $r_{\mathrm{ext},t}$ is the native game reward, and $r_{\mathrm{eth},t} \in \{0, -\rho\}$ is an ethical penalty that is nonzero iff an own-dilemma violation occurs at step $t$, with $\rho = 10$. Every utility function then refers to $\mathbf{G}_\tau$, the episode's (undiscounted) vector return:
\begin{equation}
    \mathbf{G}_\tau \;=\; \big[\, R_{\mathrm{ext}},\; R_{\mathrm{eth}} \,\big] \;=\; \sum_{t=0}^{T-1} \mathbf{r}_t .
    \label{eq:vector-return}
\end{equation}
A utility $u : \mathbb{R}^2 \to \mathbb{R}$ turns the vector into one optimizable number, and the two optimization criteria differ in where the expectation sits \citep{roijers2013survey, hayes2022practical}:
\begin{equation}
    V^{\pi}_{\mathrm{SER}} \;=\; u\big( \mathbb{E}_{\tau \sim \pi}[\mathbf{G}_\tau] \big),
    \qquad
    V^{\pi}_{\mathrm{ESR}} \;=\; \mathbb{E}_{\tau \sim \pi}\big[ u(\mathbf{G}_\tau) \big].
    \label{eq:ser-esr}
\end{equation}
For a linear $u(\mathbf{g}) = \mathbf{w}^{\top}\mathbf{g}$ the two coincide by linearity of expectation. They can differ only for non-linear $u$. SER permits cross-episode compensation (a bad episode offset by good ones), while ESR judges every episode on its own totals. Two structural facts shape the experiments. First, for any fixed $\mathbf{w} \succeq 0$ the maximizers of $\mathbf{w}^{\top} \mathbb{E}[\mathbf{G}_\tau]$ lie on the convex hull of the achievable expected-return set, so a weight sweep can only recover hull points \citep{roijers2013survey, vamplew2011empirical}, though non-linear scalarization can express more \citep{van2014multi}. Second, and decisive for how results must be read, \emph{any} statistic that is linear in the state-action occupancy measure, including the mean return-versus-violation frontier itself, cannot separate a per-episode objective from an in-expectation one. The separation, if it exists, must appear in higher moments of the per-episode distribution. Our evaluation is designed around this point.

\subsection{Dilemmas and the Ethical Penalty Layer}
We work in Craftax \citep{matthews2024craftax}, where the agent maximizes the native game reward (achievement scores plus health shaping). On top of it we define three dilemmas, each a situation where the unethical action genuinely earns more score. 

In \textbf{No Killing}, passive creatures (cows, bats, snails) are easy food and a violation is killing one, and farming (placing and eating plants) is a viable ethical alternative, so a high-return ethical policy exists. 

In \textbf{Sustainability}, wood is valuable, and each unit harvested past a per-episode budget of 8 (below the unconstrained agent's typical harvest, so the constraint binds) is a violation. 

In \textbf{Proportional Force}, hunting mobs to clear the area is easy, but a violation is killing a hostile creature (melee or ranged) that was more than 5 tiles away before the step, while closer kills count as defensive.

The ethical layer is an event-triggered penalty wrapper. A JAX-native detector classifies each step, and the violating event of the run's own dilemma receives a fixed $-10$ penalty (no positive shaping; each configuration touches only its own dilemma), so the native reward stays dominant and ethics enters as a margin, not a competing reward stream to game. Violations are detected from creature-death state differences (Section~\ref{sec:eval}), not Craftax's achievement flags, which censor repeats. For scalar training the reward is $r_t = r_{\mathrm{ext},t} + r_{\mathrm{eth},t}$ under one of three enforcement mechanisms, mechanical rather than philosophical. 

\textbf{Absolute} terminates at the first violation, \textbf{calculated} terminates once the cumulative penalty passes $-30$ (three are tolerated, the fourth ends it), and \textbf{heuristic} penalizes but continues. The wrapper carries a reward-machine-style automaton whose state never affects rewards here. We describe the layer as an event-triggered penalty and nothing more.

\subsection{Linear Baselines: SER Weight Sweep and Lagrangian Constraint}
\textbf{SER (linear MORL).} The reward is the vector $[r_{\mathrm{ext}}, r_{\mathrm{eth}}]$, scalarized as $w_{\mathrm{ext}} r_{\mathrm{ext}} + w_{\mathrm{eth}} r_{\mathrm{eth}}$ before the advantage computation, so a single scalar critic suffices \citep{roijers2013survey}. Sweeping $w_{\mathrm{eth}}$ on a log scale $\{0.1, 0.3, 1, 3, 10\}$ traces a return-versus-violation trade-off curve. For a linear utility, SER and ESR coincide, and the sweep can only recover points on the convex hull of the achievable front \citep{vamplew2011empirical}.

\textbf{Lagrangian (Constrained MDP).} This baseline comes from the constrained tradition and uses the same vector reward. With per-step cost $c_t = \max(0, -r_{\mathrm{eth},t})/\rho$ (the violations at step $t$) and episodic cost $C_\tau = \sum_t c_t$, the CMDP is
\begin{equation}
    \max_{\pi} \;\; \mathbb{E}_{\tau \sim \pi}\!\big[ R_{\mathrm{ext}} \big]
    \quad \text{s.t.} \quad
    \mathbb{E}_{\tau \sim \pi}\!\big[ C_\tau \big] \;\le\; d,
    \label{eq:cmdp}
\end{equation}
optimized through the relaxed reward $\tilde{r}_t = r_{\mathrm{ext},t} - \lambda\, c_t$ with one dual-ascent step per policy update,
\begin{equation}
    \lambda \;\leftarrow\; \mathrm{clip}\!\big( \lambda + \eta\, (\hat{C} - d),\; 0,\; \lambda_{\max} \big),
    \label{eq:dual-ascent}
\end{equation}
where $\hat{C}$ is the empirical mean episodic cost in the update window (held when the window completes no episodes). The budget $d$ is swept over $\{0.5, 1, 2, 4\}$ violations per episode. As the principled-constraint baseline, this remains a linear combination in expectation and is subject to the same convex-hull and on-average limitations. Both baselines train on the penalty-only (heuristic) configurations where their trade-off is set by the weight or budget, but never by termination.

\subsection{A Non-Compensatory Utility over Episode Returns}
The utility is defined on the episode's accumulated pair $[R_{\mathrm{ext}}, R_{\mathrm{eth}}]$, not per step, in two variants.

\textbf{Stated (primary).} An explicit smoothed thresholded-lexicographic (TLO) form \citep{gabor1998multi}. With the episode's violation count $v = \max(0, -R_{\mathrm{eth}})/\rho$ and the logistic $\sigma(x) = (1+e^{-x})^{-1}$,
\begin{equation}
    u_{\mathrm{TLO}}\big(R_{\mathrm{ext}}, R_{\mathrm{eth}}\big)
    \;=\; \sigma\!\big(s\,(\tau_{\mathrm{tol}} - v)\big)\; R_{\mathrm{ext}} \;-\; \rho_u\, v,
    \label{eq:tlo}
\end{equation}
with sharpness $s = 12$. The gate multiplies return credit by $\approx 1$ while the episode is within tolerance and by $\approx 0$ once it is not, and every violation additionally costs $\rho_u$, making the utility non-compensatory in both directions. All moral parameters are stated, visible, and hand-chosen. The form also gives $u_{\mathrm{TLO}}(\mathbf{0}) = 0$, which the telescoping identity below requires. The zero-tolerance instance ($\tau_{\mathrm{tol}} = 0.5$, $\rho_u = 10$) encodes strict compliance. The \emph{budget family} generalizes it. Setting $\tau_{\mathrm{tol}} = k + 0.5$ tolerates $k$ violations per episode before the gate closes, with a small within-budget charge ($\rho_u = 1$, deliberately below the marginal return value of a violation on No Killing, about $2.5$, so the budget region is actually exercised), swept over $k \in \{1, 2, 4\}$ on the binding dilemmas. This is ESR's analogue of the SER weight sweep and the Lagrangian budget grid, a family of exactly stated tolerance-$k$ preferences. It carries the property the experiments test, namely that a per-episode budget is a preference that no linear, occupancy-additive objective can express.

\textbf{Distilled (secondary).} We roll out a pool of trained agents (unconstrained, the ERM agents, the SER sweep), record each episode's $[R_{\mathrm{ext}}, R_{\mathrm{eth}}]$, and label preference pairs by the ethics-gated rule above plus a monotonicity prior. A small MLP $u_\theta$ minimizes the Bradley--Terry loss \citep{bradley1952rank, christiano2017deep} over winners $\mathbf{G}_w$ and losers $\mathbf{G}_l$,
\begin{equation}
    \mathcal{L}(\theta) \;=\; -\,\mathbb{E}_{(w,l)}\Big[ \log \sigma\big( u_\theta(\mathbf{G}_w) - u_\theta(\mathbf{G}_l) \big) \Big],
    \label{eq:bt}
\end{equation}
and its output is rescaled at load time to game-return units (a positive constant, preserving the ordering and the ESR argmax). The labels are ours, so the fitted utility \emph{distills} a stated ordering. The variant tests whether fitting helps or hurts relative to stating it.

\subsection{ESR-PPO: Optimizing the Utility per Episode}
The ESR trainer maximizes $V^{\pi}_{\mathrm{ESR}} = \mathbb{E}[\,u(\mathbf{G}_\tau)\,]$ of Eq.~\eqref{eq:ser-esr}, the utility of each episode's own totals, rather than $V^{\pi}_{\mathrm{SER}} = u(\mathbb{E}[\mathbf{G}_\tau])$ \citep{roijers2020multi, ruadulescu2020multi, hayes2022practical}. Under a non-compensatory $u$, a violating episode cannot be paid for by other episodes. Two standard devices make this trainable with PPO. First, a telescoping pseudo-reward. With the within-episode accumulator $\mathbf{R}^{\mathrm{acc}}_t = \sum_{k \le t} \mathbf{r}_k$ (reset at episode start, $\mathbf{R}^{\mathrm{acc}}_{-1} = \mathbf{0}$), each step's scalar reward is the utility increment
\begin{equation}
\begin{aligned}
    r^{u}_t &\;=\; u\big(\mathbf{R}^{\mathrm{acc}}_t\big) - u\big(\mathbf{R}^{\mathrm{acc}}_{t-1}\big), \\
    \sum_{t=0}^{T-1} r^{u}_t &\;=\; u\big(\mathbf{G}_\tau\big) - u(\mathbf{0}) \;=\; u\big(\mathbf{G}_\tau\big).
\end{aligned}
    \label{eq:telescope}
\end{equation}
so standard PPO/GAE on $r^{u}$ optimizes the ESR objective. The identity is exact at $\gamma = 1$ but for $\gamma < 1$ it introduces a bias we quantify with a $\gamma = 1$ arm. Second, since the ESR-optimal action depends on what has accumulated (the distance to the gate makes the criterion non-stationary in the plain state), the observation is extended to $\tilde{\mathbf{o}}_t = [\, \mathbf{o}_t \,;\; \nu(\mathbf{R}^{\mathrm{acc}}_t) \,]$ with $\nu$ as a fixed normalizer. Both devices follow the accrued-return ESR line of \citet{reymond2023actor}. They are the vehicle, not the contribution. The augmentation is a potential confound, so we also run \textbf{SER$+R_{\mathrm{acc}}$}, which is an identical setup with a \emph{linear} utility, whose increments collapse to the per-step SER reward $\mathbf{w}^{\top}\mathbf{r}_t$, i.e.\ the SER objective under the ESR observation. If ESR wins only through the extra input, this control reproduces the win. If it does not, the objective is responsible. Everything else (network, hyperparameters, environment stack) is identical across methods, isolating the objective.

\section{Evaluation Methodology}
\label{sec:eval}
Every trained agent, however trained, is rolled out in the same detector-only environment (the detector runs but adds no reward and never terminates). The sustainability budget is read from each run's training config so all face the same threshold.

\textbf{Uncensored counting.} Violations are counted from creature-death state differences, with per-step counts (a step where two creatures are killed, counts as two). The alternative, Craftax's achievement flags, latches once per creature type per episode which would understate unsafe agents and make per-episode dispersion a deterministic function of the mean ($\mathrm{std} = \sqrt{m(1-m)}$ for a near-binary count), voiding the distributional comparison. The detector is unit-tested both ways, so that a never-attacking policy registers zero kills and an always-attacking one registers many.

\textbf{Distributions and not only means.} For per-episode violation counts $\{V_i\}_{i=1}^N$ we report the mean and standard deviation, the violating fraction $P(\geq 1) = \tfrac{1}{N}\sum_i \mathbf{1}[V_i \ge 1]$, and the tail mean $\mathrm{CVaR}_{0.1}$ (mean of the worst $\lceil 0.1 N \rceil$ episodes). The main metric is the \emph{matched-return tail comparison}. Each ESR point is paired with the return-nearest SER weight, Lagrangian budget, and SER$+R_{\mathrm{acc}}$ control point on the same dilemma, and we compare distributions at equal mean returns. This is the only place where a tail difference is attributable to the objective rather than to a position on the curve. We also report the frontier, the cost of ethics $\mathrm{gap}(m) = \bar{R}_{\mathrm{uncon}} - \bar{R}_{m}$ with a variance-propagated 95\% CI over $n = 3$ seeds, and hypervolume, but the $\gamma = 1$ component is reported separately.

\section{Experimental Setup}
All experiments use Craftax-Symbolic-v1 \citep{matthews2024craftax}. Episodes end by in-game death, by timeout, or, during ERM training only, by the enforcement mechanism. Every method trains the same agent, PPO with a GRU recurrent actor-critic (hidden size 512), 1024 parallel environments, $10^9$ environment steps per run, Adam with linear learning-rate decay, $\gamma = 0.99$, and GAE $\lambda = 0.8$ (full hyperparameter table in the supplementary material). The run matrix uses 3 seeds throughout. It covers ERM at 3 dilemmas $\times$ 3 enforcements plus unconstrained (30 runs), SER at 3 dilemmas $\times$ 5 weights (45), Lagrangian at 3 dilemmas $\times$ 4 budgets (36), and ESR at 3 dilemmas $\times$ \{stated, distilled\} (18). To these we add the budget arms on the two binding dilemmas, $k \in \{1,2,4\}$ (18), the SER$+R_{\mathrm{acc}}$ control at two matched weights (18), and the $\gamma = 1$ twins (18), for 183 runs in total. All agents are scored by one evaluation job of 64 environments $\times$ 4096 steps per agent ($\approx 2{,}400$ completed episodes per operating point across seeds). Each training run takes $\approx 3$ hours on a single H100 GPU ($\approx 92$k steps/s) on national HPC clusters. The ESR stage was retrained from scratch on a second cluster after a mid-study migration and reproduced its results.

\section{Results}

\subsection{Mean Return and Violation Rates}
\begin{figure}[t]
\centering
\includegraphics[width=\columnwidth]{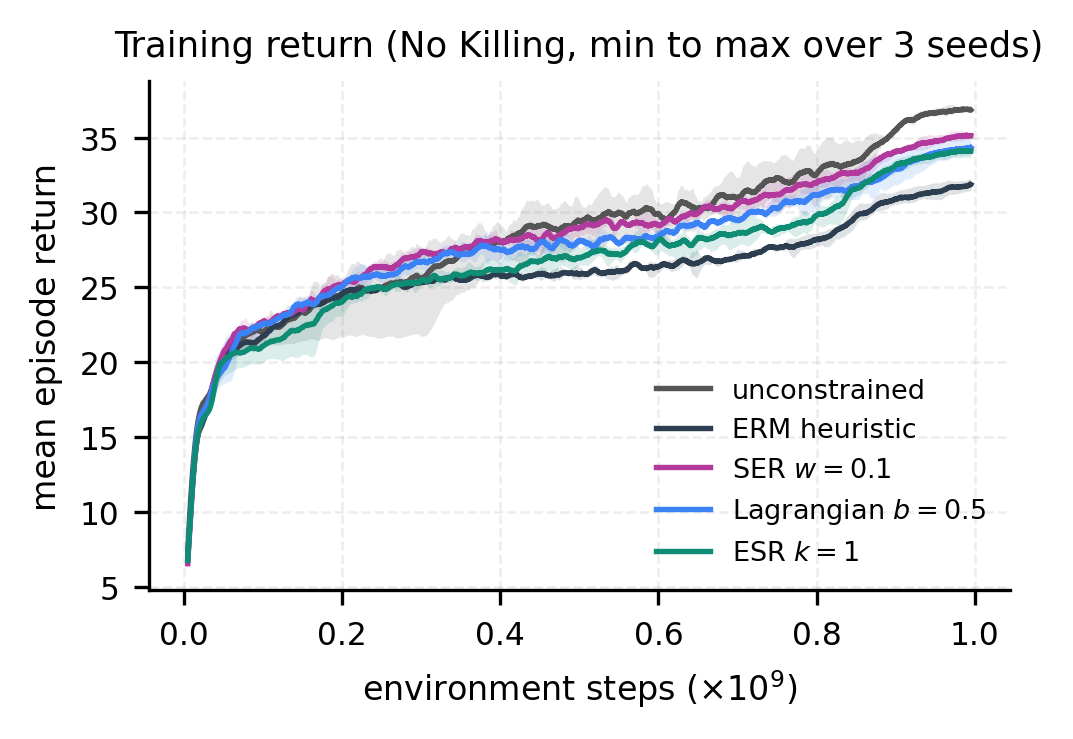}
\caption{Training return on No Killing (mean with min--max band over 3 seeds). Every method family converges with the same shape, and the compliant methods plateau together about 6 points below the unconstrained agent, so the shared frontier of Figure~\ref{fig:frontier} reflects the cost of compliance rather than an under-tuned baseline.}
\label{fig:training}
\end{figure}

\begin{figure}[t]
\centering
\includegraphics[width=\columnwidth]{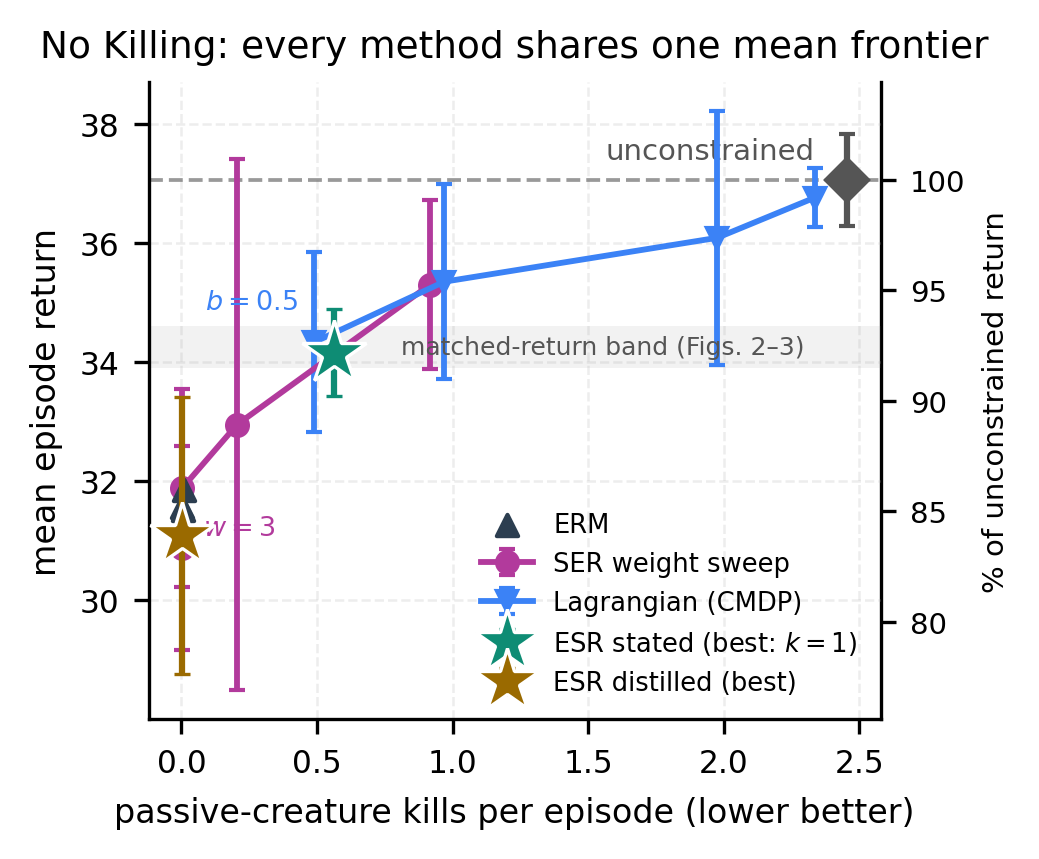}
\caption{No Killing frontier in per-episode units (mean $\pm$ 95\% CI, 3 seeds, right axis \% of unconstrained return). The SER weight sweep and Lagrangian budgets trace one shared envelope, and the best point of each ESR version (stated budget $k{=}1$ and distilled) lands on it. The shaded band marks the matched-return comparison of Figures~\ref{fig:distribution}--\ref{fig:budget}. The full ESR family is Figure~\ref{fig:esr_variations}, and the second binding dilemma (Sustainability) is in the supplementary material.}
\label{fig:frontier}
\end{figure}

As intended, all four methods reach a common return-versus-violation frontier, so the aggregate view cannot separate them. Every method trains stably, with final returns between 30 and 37 (Figure~\ref{fig:training}). The unconstrained agent scores $37.05 \pm 0.78$ and is genuinely tempted, killing a passive creature in 95\% of No-Killing episodes and over-harvesting about 5 units per Sustainability episode. Driving violations to near zero costs every method roughly the same amount. Mean return falls to 31.5--31.9 for ERM, 30.9--31.0 for the strictest SER weights, and 30.9--31.1 for both ESR variants, in each case about 84\% of the unconstrained return. This 5--6 point gap is the price of compliance in this environment rather than a property of any single method. The pattern holds along the whole frontier (Figure~\ref{fig:frontier}). The ESR budget points lie on the SER and Lagrangian envelope rather than above it, and adding ESR to the coverage set raises hypervolume only slightly (0.0264 against 0.0249 on No Killing). This is what our problem setting anticipates, and it is the outcome we want. The mean return-versus-violation frontier is linear in the state-action occupancy measure, so every point on it is attainable by some linear weighting, and a per-episode objective cannot distinguish itself there. A method that appeared to beat the linear baselines here would point to under-tuned baselines rather than to a better objective. Any difference between the methods must instead appear in how violations are distributed across episodes, which is what the matched-return comparison below is designed to expose.

\subsection{Per-Episode Violation Distributions}
\begin{figure}[t]
\centering
\includegraphics[width=\columnwidth]{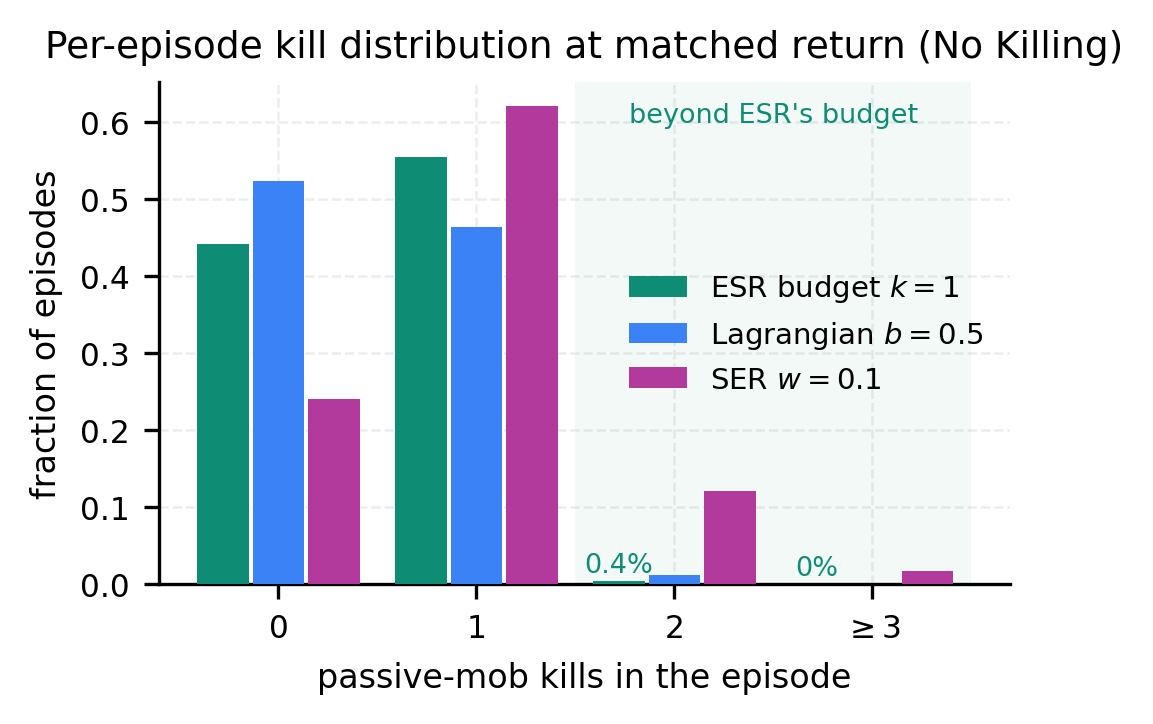}
\caption{Per-episode violation-count distributions for the three return-matched agents on No Killing (about 2{,}400 episodes each). The ESR budget agent's mass stops at its one-violation budget, with 0.4\% of episodes at two kills and none beyond. The Lagrangian reaches three, and the matched SER weight develops a longer tail (12.1\% of episodes at two kills and 1.7\% at three or more).}
\label{fig:distribution}
\end{figure}

The methods separate clearly once we examine how violations are distributed across episodes rather than only their average. We compare three agents at closely matched mean return on No Killing and read off their per-episode violation counts (Figure~\ref{fig:distribution}, Table~\ref{tab:claimb}). The ESR budget agent and the Lagrangian sit at 34.2 and 34.3, and the return-nearest SER weight sits a little higher at 35.3, so the comparison is if anything generous to SER. The ESR budget agent ($k{=}1$) stays within its stated budget of one violation per episode. Across roughly 2{,}400 episodes it exceeds the budget in 0.4\% of them and never records three violations, and its worst-decile mean is $\mathrm{CVaR}_{10\%} = 1.04 \pm 0.07$. The Lagrangian ($b{=}0.5$) meets the same budget on average but not within each episode. In 1.3\% of its episodes it exceeds one violation, and its worst decile averages $1.14 \pm 0.03$ (Welch $t \approx 5.5$ against ESR). The return-nearest SER weight ($w{=}0.1$) has a heavier tail again, with 13.8\% of episodes over budget and a worst decile of $2.20 \pm 0.20$ ($t \approx 23$), twice the ESR agent's. The SER$+R_{\mathrm{acc}}$ control isolates the cause. It gives the linear objective the same accrued-return observation as ESR, yet it stays on the SER frontier rather than reaching ESR's operating point. At its own return of 32.5 its tail matches the nearby SER weight (CVaR 1.12 against 1.18 for plain SER at 32.9), not ESR's. The tighter tail therefore comes from the objective, not from the extra input.

\begin{figure}[t]
\centering
\includegraphics[width=\columnwidth]{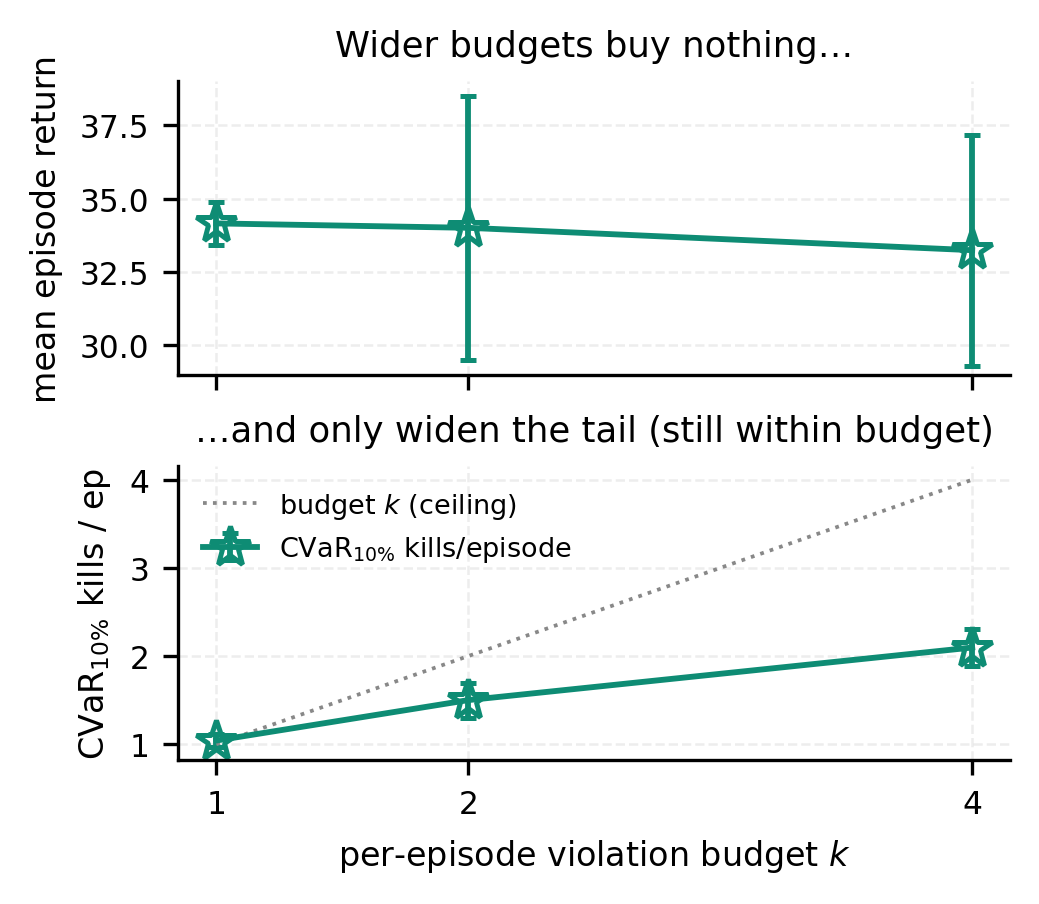}
\caption{Budget sweep on No Killing. Widening the tolerance $k = 1 \rightarrow 2 \rightarrow 4$ buys no return (top) and only widens the worst-decile tail toward, but not past, the budget ceiling (bottom). The temptation is concentrated in the first violation.}
\label{fig:budget}
\end{figure}

Widening the tolerance does not change the picture (Figure~\ref{fig:budget}). Raising $k$ from 1 to 2 to 4 leaves the mean return essentially flat (34.15, 34.00, 33.24) while the worst-decile tail grows toward, but never past, each new ceiling (CVaR 1.04, 1.50, 2.10). In this dilemma, the temptation is concentrated in the first violation. The same ordering holds on Sustainability more sharply, there the agents match at about 34.5 mean return. The ESR budget agent's worst-decile tail is $\mathrm{CVaR}_{10\%} = 1.14$ against the return-nearest Lagrangian's 3.73, and it exceeds one violation in 10\% of episodes against the Lagrangian's 55\%. Sustainability is a weaker test in other respects, because our single within-budget charge of $\rho_u = 1$ sat close to the marginal return value of a violation there, about 1.1 against about 2.5 on No Killing, leaving little incentive to spend the budget, which was therefore only lightly exercised. 

A single strong SER weight dominates the region on the mean, and one seed collapses, but the per-episode gap runs in the same direction as on No Killing. Stated precisely, the objective removes cross-episode compensation and, on No Killing, holds violations within budget in 99.6\% of episodes. It does not make any individual episode certain to be clean, since the policy and environment remain stochastic.
\begin{figure}[t]
\centering
\includegraphics[width=\columnwidth]{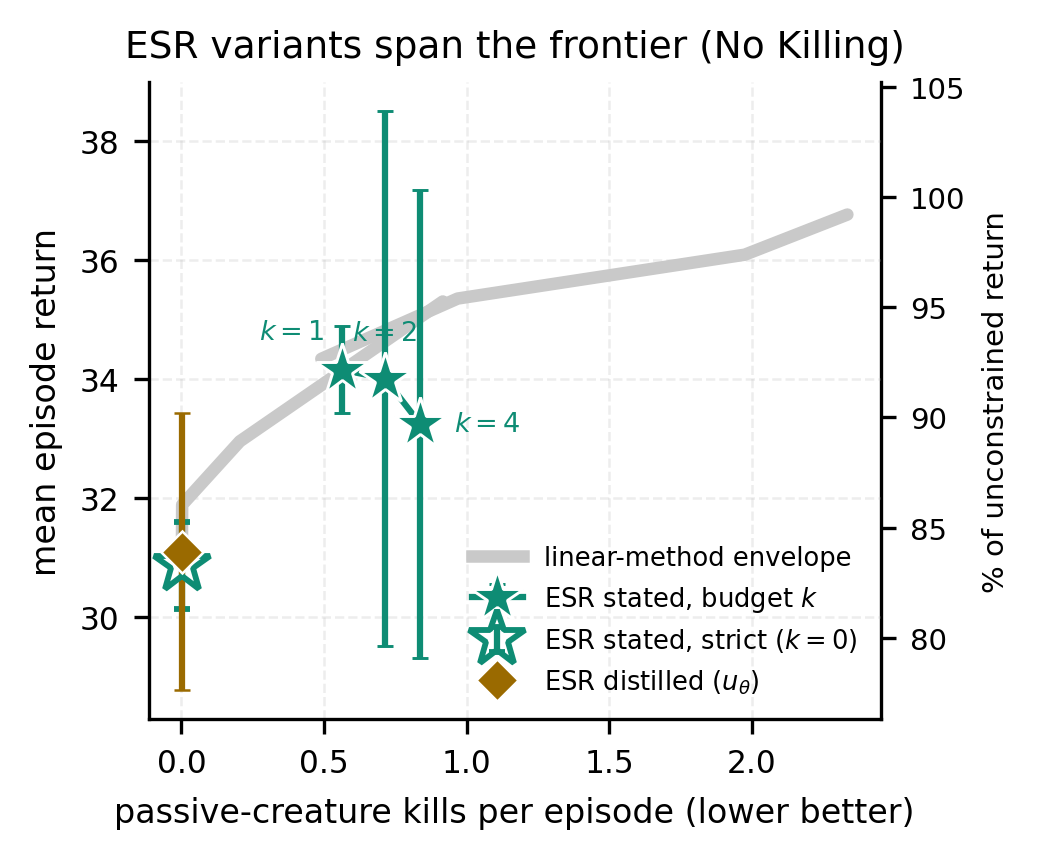}
\caption{The ESR family on the No-Killing frontier, with the SER/Lagrangian envelope in gray for reference. The stated budget sweep ($k = 1, 2, 4$) climbs the envelope, and the strict stated point ($k{=}0$) and the distilled utility $u_\theta$ sit together at the compliance corner. The variants choose \emph{where} on the shared frontier to operate.}
\label{fig:esr_variations}
\end{figure}

\begin{table}[t]
\centering
\caption{Per-episode violation statistics for the three agents compared at closely matched mean return on No Killing. The return-nearest SER weight sits a little higher at 35.3, so the comparison favors SER, yet the non-compensatory ESR objective still holds the tightest tail. The SER$+R_{\mathrm{acc}}$ control is discussed in the text, and complete tables for every method and dilemma are released with the code.}
\label{tab:claimb}
\small
\setlength{\tabcolsep}{5pt}
\begin{tabular}{lrrrr}
\toprule
Method (No Killing) & Return & $P(\geq 1)$ & Std & CVaR$_{10\%}$ \\
\midrule
ESR budget ($k{=}1$)     & 34.2 & 0.56 & 0.50 & \textbf{1.04} \\
Lagrangian ($b{=}0.5$)   & 34.3 & 0.48 & 0.53 & 1.14 \\
SER ($w{=}0.1$)          & 35.3 & 0.76 & 0.66 & 2.20 \\
\bottomrule
\end{tabular}
\end{table}

\subsection{The Cost of Compliance}
A tighter tail might be expected to cost return, but it does not. At the strict-compliance corner every method loses between 5.2 and 6.2 return relative to the unconstrained agent (about 16\%), and at a one-violation budget the ESR and Lagrangian agents lose statistically the same amount ($2.90 \pm 1.06$ and $2.72 \pm 1.70$, Table~\ref{tab:gap}). The per-episode guarantee therefore adds no return penalty over the linear baselines. Proportional Force is a limitation of the benchmark. Our detector shows that the constraint does not bind, since the unconstrained agent strikes preemptively only $1.4 \times 10^{-5}$ of the time per step. There the distilled utility, fitted to a pool whose worst episode contains a single violation, loses about 1.5 return for no gain in safety, the failure mode that our stated-versus-distilled comparison is designed to reveal (supplementary material).
\begin{table}[t]
\centering
\caption{Cost of compliance on No Killing. We report mean return, own-dilemma violation rate, $P(\geq 1)$, and gap to the unconstrained agent (means over three seeds). Every compliant method pays a similar 5--6 point gap at the strict corner, and the ESR budget arm and the Lagrangian pay the same to operate at a one-violation budget. Full tables for all methods and dilemmas are released with the code.}
\label{tab:gap}
\small
\setlength{\tabcolsep}{5pt}
\begin{tabular}{lrrrr}
\toprule
Method (No Killing) & Return & Viol rate & $P(\geq 1)$ & Gap $\downarrow$ \\
\midrule
Unconstrained            & 37.1 & 7.3e-03 & 0.95 & 0.0 \\
ERM (heuristic)          & 31.8 & 2.3e-05 & 0.01 & 5.2 \\
SER ($w{=}0.1$)          & 35.3 & 3.0e-03 & 0.76 & 1.8 \\
Lagrangian ($b{=}0.5$)   & 34.3 & 1.6e-03 & 0.48 & 2.7 \\
ESR strict ($k{=}0$)     & 30.9 & 6.4e-06 & 0.00 & 6.2 \\
ESR budget ($k{=}1$)     & 34.2 & 1.7e-03 & 0.56 & 2.9 \\
\bottomrule
\end{tabular}
\end{table}

\subsection{Ablations and Robustness Checks}
\textbf{ESR variants.} Figure~\ref{fig:esr_variations} places the full ESR family on the No-Killing frontier. The stated budget sweep runs along the frontier from the compliance corner upward, and the strict stated point ($k{=}0$) and the distilled utility coincide at the corner. The variants differ in where on the frontier they operate, not in whether they reach it. Adding ESR to the coverage set raises hypervolume only slightly ($+6\%$ on No Killing), consistent with the shared frontier.

\textbf{Stated versus distilled utility.} The two match at the corner (30.9 against 31.1 return), but the distilled utility is noisier wherever its training pool held few violations. On Proportional Force it scores $35.2 \pm 4.5$ against the stated utility's $36.7 \pm 1.1$ (utility surfaces in the supplementary material). Fitting the utility adds nothing over stating it and can cost return. 

\textbf{Discounting.} Comparing $\gamma = 1$ with $\gamma = 0.99$ shows that the telescoping approximation introduces a real but dilemma-dependent bias, costing about 3 return on No Killing at held compliance while improving Sustainability. We report it as a sensitivity check, and the No-Killing result is stable across both settings.

\textbf{Reproducibility.} Every method family converges without collapse. The ESR runs reproduced their corner returns to within $\pm 0.3$ when retrained from scratch on different hardware.

\section{Discussion}
Every method that optimizes expected returns shares one blind spot. A bad episode can be offset by good ones. Whether that trade is available is decided by the shape of the utility, not by the training algorithm. This gives our results a two-sided reading. Judged on mean return and mean violation rate, as most multi-objective and constrained RL is judged, the four methods are interchangeable and the Lagrangian is the simplest good choice. Judged per episode, the three expectation-level methods spread their violations unevenly across episodes, which for an ethical constraint is the behavior that matters. The per-episode evaluation is therefore not a secondary diagnostic. It is what makes the effect visible. In practical terms, a per-episode budget utility is the right tool when compliance must hold within each episode, an expected-cost constraint is enough when only long-run rates matter, and a utility should be distilled from data only when that data contains violations to learn from.

\paragraph{Relation to risk-sensitive RL.}
Because we report CVaR, our approach may be read as risk-sensitive RL, but the two act on different quantities. Risk-sensitive methods optimize a tail statistic, a CVaR or a chance constraint, of the \emph{return} distribution, reshaping the spread of rewards, and they take the risk level as a training input. We instead make the per-episode ethics budget the argument of the utility and use CVaR only to evaluate, so the tail we control is the tail of the \emph{violation} count rather than of return. The two are therefore complementary rather than competing, and a CVaR-constrained agent would be a natural additional baseline.

\paragraph{Scope.}
We do not claim that RL should abandon mean-based evaluation in general. The argument applies when violations do not compensate across episodes. Ethical harms are of this kind, since a harm in one episode is not undone by good conduct in another, and their natural target is the per-episode distribution that a mean averages away. When a constraint really is about long-run behavior, such as a power or bandwidth budget or an average-cost target, violations are fungible across episodes, the mean is the correct target, and a simpler in-expectation method is appropriate. Our claim is that in the non-compensatory case both the training objective and the evaluation statistic should move from the expectation to the distribution. We demonstrate this in one environment and expect it to hold wherever that structure is present, rather than for RL evaluation as a whole.

\paragraph{Limitations.}
Our proportionality constraint is defined by distance alone, so an agent that kills a distant ranged attacker is counted as striking preemptively, and the dilemma captures distance rather than threat. The training penalty is applied per step and slightly under-penalizes the rare step that removes two creatures at once, although measurement uses exact counts. Separately, a kill is missed if a new creature spawns into the same slot within the step. The telescoping identity is exact only at $\gamma = 1$. At $\gamma = 0.99$ it introduces a bounded bias of up to 3--4 return that we measure rather than remove, shared across all ESR runs. Finally, we use three seeds, so intervals are wide. The main tail comparisons are significant (Welch $t \approx 5.5$ and $23$), but one Sustainability seed collapsed, and all results come from a single environment with distillation labels generated by our own rule.

\section{Conclusion and Future Work}
We compared four ways of training ethical behavior across three Craftax dilemmas under a single per-episode evaluation protocol. The four methods are indistinguishable on the mean return-versus-violation frontier, but per episode only the non-compensatory objective keeps its stated violation budget in effectively every episode, at no cost in mean return, and a control attributes this to the objective rather than to the agent's observation. The broader point concerns measurement as much as training. When violations cannot be undone across episodes, an agent that is ethical on average is not ethical, and an evaluation that reports only mean rates cannot see the difference. Future work includes preferences from human raters, social-dilemma environments in which the penalty layer maps onto native metrics, and a Saute-style hard-budget baseline \citep{sootla2022saute}.

\bibliography{aaai2027}

\appendix
\section*{Supplementary Material}
% Content extracted from appendix.tex for inclusion at the end of arxiv.tex

\section{Hyperparameters}
Table~\ref{tab:hyper} lists the PPO hyperparameters used for every run in the study. They are held fixed across all methods and dilemmas so that the comparison isolates the training objective.

\begin{table}[h]
\centering
\caption{Training hyperparameters, identical for every method.}
\label{tab:hyper}
\small
\begin{tabular}{ll}
\toprule
Hyperparameter & Value \\
\midrule
Recurrent core & GRU, hidden size 512 \\
Parallel environments & 1024 \\
Rollout length & 64 steps \\
Environment steps per run & $10^9$ \\
Optimizer & Adam ($\epsilon = 10^{-5}$) \\
Learning rate & $2 \times 10^{-4}$, linear decay \\
Discount $\gamma$ & 0.99 \\
GAE $\lambda$ & 0.8 \\
PPO clip $\epsilon$ & 0.2 \\
Update epochs & 4 \\
Minibatches & 8 \\
Entropy coefficient & 0.01 \\
Value coefficient & 0.5 \\
Max gradient norm & 1.0 \\
Seeds & 42, 123, 7 \\
\bottomrule
\end{tabular}
\end{table}

\section{Additional Figures}

\begin{figure*}[t]
\centering
\includegraphics[width=\textwidth]{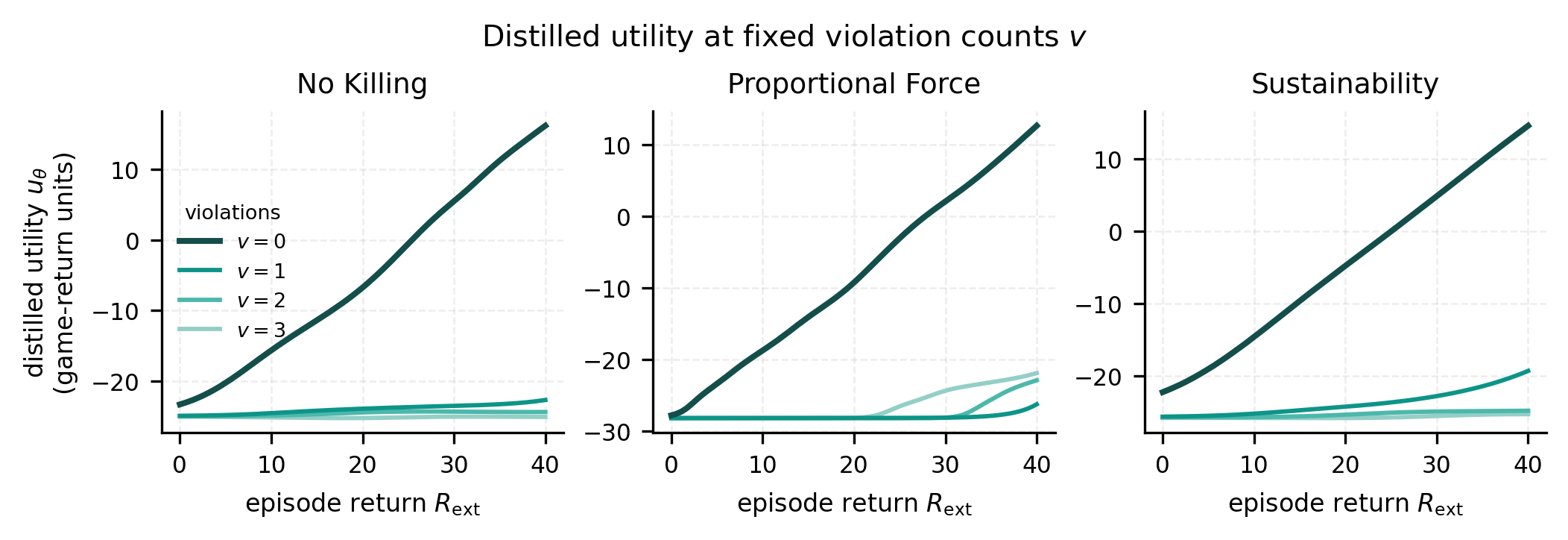}
\caption{Distilled utility sliced at fixed violation counts $v$, in game-return units. On No Killing and Sustainability the fit is well behaved: a sharp cliff separates $v = 0$ from $v \geq 1$, and utility decreases monotonically with further violations. On Proportional Force the cliff survives but the ordering among violating episodes inverts at high return, where the fit scores three violations above one.}
\label{fig:sm_utility}
\end{figure*}

\begin{figure}[t]
\centering
\includegraphics[width=\columnwidth]{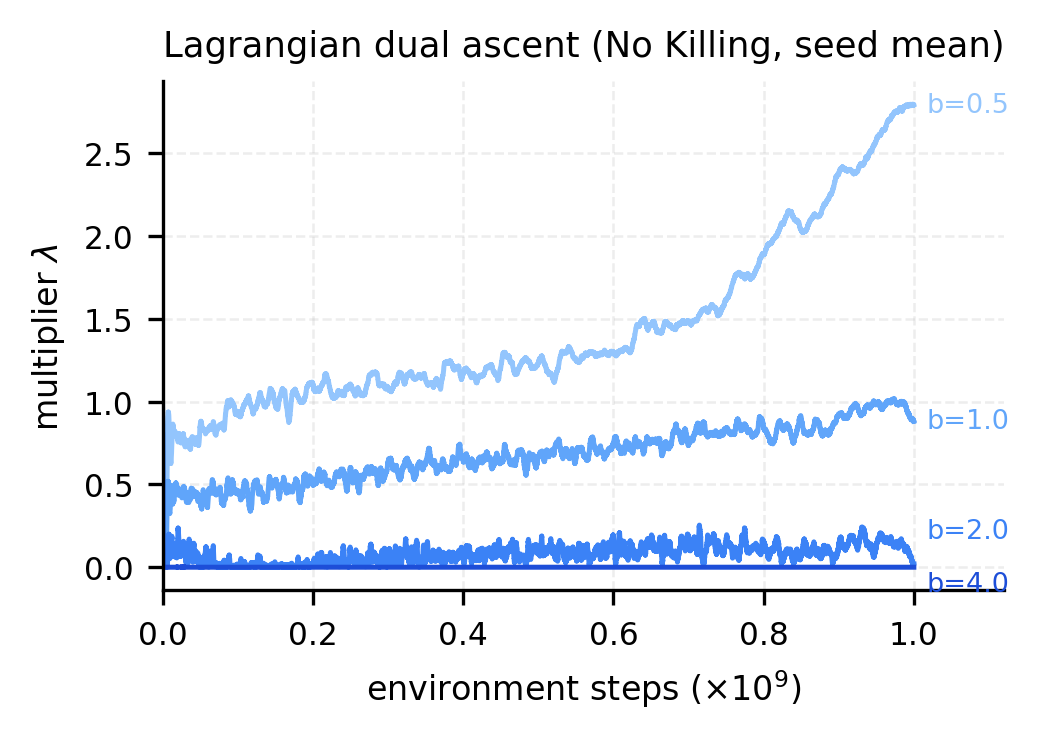}
\caption{Lagrangian dual ascent on No Killing (seed mean). The multiplier $\lambda$ rises only for budgets tighter than the unconstrained agent's violation rate, and stays near zero for the loose budgets that never bind, as designed.}
\label{fig:sm_curves}
\end{figure}

\begin{figure*}[t]
\centering
\includegraphics[width=\textwidth]{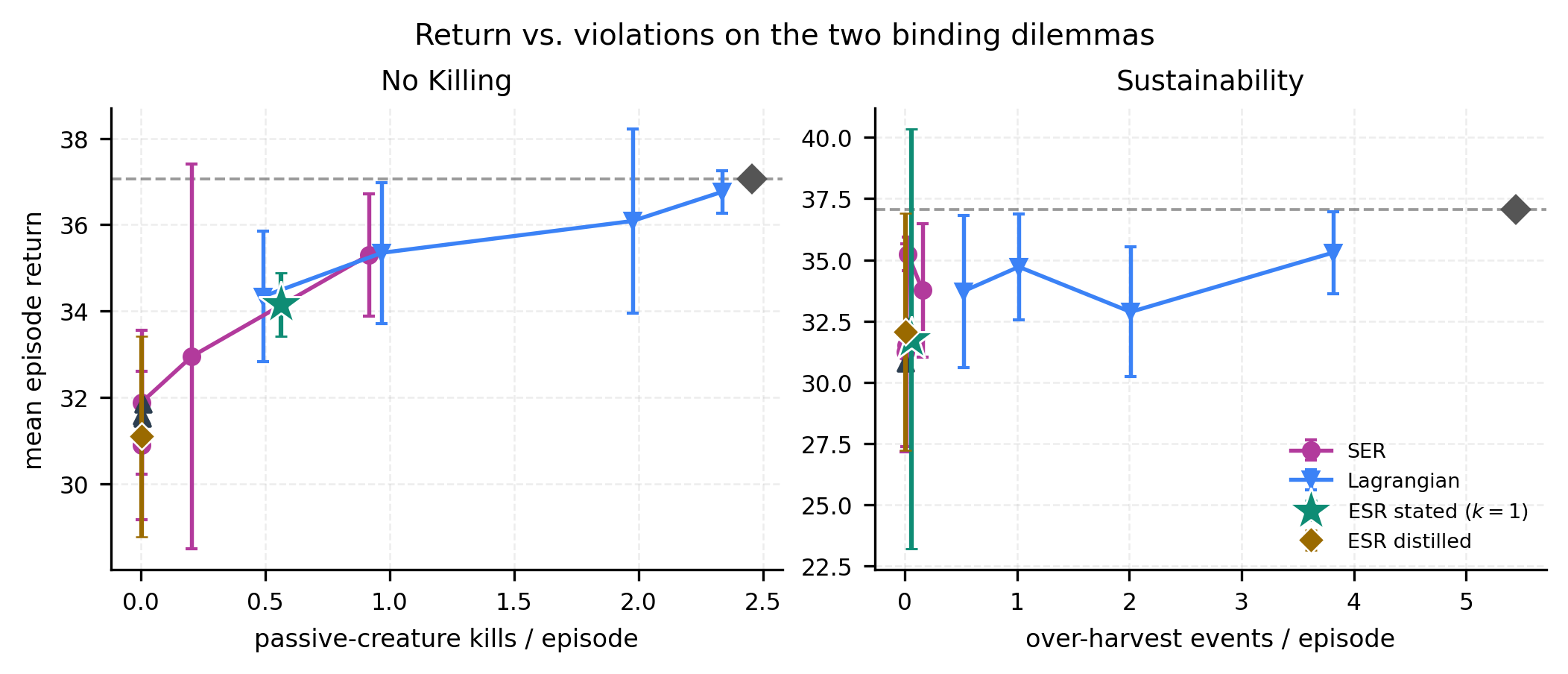}
\caption{Return vs.\ per-episode violations on the two binding dilemmas (mean $\pm$ 95\% CI, 3 seeds). No Killing (left) reproduces the main-paper frontier in full; Sustainability (right) shows the same shared-envelope pattern, although its ESR budget arms were under-exercised. Proportional Force is omitted as non-binding.}
\label{fig:sm_frontier}
\end{figure*}

\paragraph{Frontiers on the binding dilemmas.}
The main paper reports the No Killing frontier in per-episode units. Figure~\ref{fig:sm_frontier} adds Sustainability, the second dilemma that binds; it shows the same shared-envelope pattern, although the ESR budget arms there were only lightly exercised (see the Cost of Compliance section of the main paper). Proportional Force is omitted because it does not bind under state-diff detection: the unconstrained agent's preemptive-strike rate is $1.4 \times 10^{-5}$ per step, so a return-versus-violation frontier there carries no signal, and all methods collapse to the same near-zero-violation corner. Its role is instead the distillation case study of Figure~\ref{fig:sm_utility}. Figure~\ref{fig:sm_curves} shows the Lagrangian dual ascent; training convergence is reported in the main paper.

\paragraph{Why the distilled utility degrades on a violation-free pool.}
The distilled utility is fit to the return pairs the exemplar agents actually realise, so it is constrained only where those agents produced data. Figure~\ref{fig:sm_utility} slices the fitted utility at the violation counts a policy visits. The No Killing and Sustainability pools contain episodes with up to 11 and 26 violations, and both fits behave as intended: a sharp cliff separates clean episodes from violating ones, and utility falls monotonically as violations accumulate. The Proportional Force pool is different. Its worst single episode contains one violation, so the fitted $R_{\mathrm{eth}}$ axis spans only $[-10, 0]$ and nothing constrains the fit beyond a single violation. The cliff at the first violation still forms, because the pool does contain both clean and violating episodes, but the ordering among violating episodes inverts: at $R_{\mathrm{ext}} = 40$ the fitted utility assigns $-21.9$ to three violations against $-26.2$ to one, ranking the worse episode higher. An agent trained on this utility receives incoherent guidance as soon as it violates at all, which is why it pays about 1.5 return for no gain in safety. This is the intended function of the stated-versus-distilled ablation: it flags exactly the regime, a pool that does not exercise the ethical axis, where distillation should not be trusted.

\end{document}